\documentclass{article}

     \usepackage[final]{neurips_2020}

\usepackage[utf8]{inputenc} 
\usepackage[T1]{fontenc}    
\usepackage{hyperref}       
\usepackage{url}            
\usepackage{booktabs}       
\usepackage{amsfonts}       
\usepackage{nicefrac}       
\usepackage{microtype}      
\usepackage{graphicx}

\title{A Survey of Deep Learning Approaches for OCR and Document Understanding}

\author{%
  Nishant Subramani\\
  Intel Labs\\
  Masakhane\\
  \texttt{nishant.subramani23@gmail.com}\\
  \And
  Alexandre Matton\\
  Scale AI\\
  \texttt{alexandre.matton@scale.com}\\
  \AND
  Malcolm Greaves\\
  Scale AI\\
  \texttt{malcolm.greaves@scale.com}\\
  \And
  Adrian Lam\\
  Scale AI\\
  \texttt{adrian.lam@scale.com}\\
}

\begin{document}

\maketitle

\begin{abstract}
Documents are a core part of many businesses in many fields such as law, finance, and technology among others.
Automatic understanding of documents such as invoices, contracts, and resumes is lucrative, opening up many new avenues of business.
The fields of natural language processing and computer vision have seen tremendous progress through the development of deep learning such that these methods have started to become infused in contemporary document understanding systems.
In this survey paper, we review different techniques for document understanding for documents written in English and consolidate methodologies present in literature to act as a jumping-off point for researchers exploring this area. 
\end{abstract}

\section{Introduction}
Humans compose documents to record and preserve information. As information carrying vehicles, documents are written using different layouts to represent diverse sets of information for a variety of different consumers.
In this work, we look at the problem of document understanding for documents written in English. Here, we take the term document understanding to mean the automated process of reading, interpreting, and extracting information from the written text and illustrated figures contained within a document's pages. From the perspective as practitioners of machine learning, this survey covers the methods by which we build models to automatically understand documents that were originally composed for human consumption.
Document understanding models take in documents and segment pages of documents into useful parts (i.e. regions corresponding to a specific table or property), often using optical character recognition (OCR)~\citep{mori1999optical} with some level of document layout analysis.
These models use this information to understand the contents of the document at large, e.g. that this region or bounding box corresponds to an address.
In this survey, we focus on these aspects of document understanding at a more granular level and discuss popular methods for these tasks.
Our goal is to summarize the approaches present in modern document understanding and highlight current trends and limitations.

In Section~\ref{sec:document-processing-understanding}, we discuss some general themes in modern NLP and document understanding and provide a framework for building end-to-end automated document understanding systems.
Next, in Section~\ref{sec:ocr}, we look at the best methods for OCR encompassing both text detection (Section~\ref{sec:text-detection}) and text transcription (Section~\ref{sec:text-transcription}).
We take a broader view of the document understanding problem in section~\ref{sec:layout-analysis}, presenting multiple approaches to document layout analysis: the problem of locating relevant information on each page.
Following this, we discuss popular approaches for information extraction (Section~\ref{sec:information-extraction}).

\section{Document Processing \& Understanding}\label{sec:document-processing-understanding}
Document processing historically involved handcrafted rule-based algorithms~\citep{layout201771, Ha1995DocumentPD, Amin2001PageSA}, but with the widespread success of deep learning~\citep{collobert2008unified, krizhevsky2012imagenet, sutskever2014sequence}, computer vision (CV) and natural language processing (NLP) based methods have come to the fore.
Advancements in object detection and image segmentation have led to systems that edge close to human performance on a variety of tasks~\citep{redmon2016you, lin2017feature}.
As a result, these methods have been applied to a variety of other domains including NLP and speech~\citep{Gehring2017ConvolutionalST, Wu2018ALC, Subramani2020LearningER}.
Since documents can be read and viewed as a visual information medium, many practitioners leverage computer vision techniques as well and use them for text detection and instance segmentation~\citep{long2018textsnake,Katti2018ChargridTU}. 
We cover specific methods to do these in Sections~\ref{sec:text-detection} and~\ref{sec:instance-segmentation}.

The widespread success and popularity of large pretrained language models such as ELMo and BERT have caused document understanding to shift towards using deep learning based models~\citep{peters2018deep, Devlin2019BERTPO}.
These models can be fine-tuned for a variety of tasks and have replaced word vectors as the de-facto standard for pretraining for natural language tasks.
However, language models, both recurrent neural network based and transformer based~\citep{vaswani2017attention}, struggle with long sequences~\citep{Cho2014OnTP, Subramani2019CanUL, Subramani2020DiscoveringUS}.
Given that texts can be very dense and long in business documents, model architecture modifications are necessary.
The most simple approach is to truncate documents into smaller sequences of 512 tokens such that pretrained language models can be used off-the-shelf~\citep{Xie2019UnsupervisedDA, Joshi2019BERTFC}. Another approach that has gained traction recently is based on reducing the complexity of the self-attention component of transformer-based language models \citep{child2019generating, Beltagy2020LongformerTL, katharopoulos2020transformers, kitaev2020reformer, choromanski2020rethinking}.

All effective, modern, end-to-end document understanding systems present in the literature integrate multiple deep neural network architectures for both reading and comprehending a document's content. Since documents are made for humans, not machines, practitioners must combine CV as well as NLP architectures into a unified solution. While specific use cases will dictate the exact techniques used, a full end-to-end system employs:
\begin{itemize}
    \item A computer-vision based document layout analysis module, which partitions each document page into distinct content regions. This model not only delineates between relevant and irrelevant regions, but also serves to categorize the type of content it identifies.
    \item An optical character recognition (OCR) model, whose purpose is to locate and faithfully transcribe all written text present in the document. Straddling the boundary between CV and NLP, OCR models may either use document layout analysis directly or solve the problem in an independent fashion.
    \item Information extraction models that use the output of OCR or document layout analysis to comprehend and identify relationships between the information that is being conveyed in the document. Usually specialized to a particular domain and task, these models provide the structure necessary to make a document machine readable, providing utility in document understanding.
\end{itemize}

In the following sections, we expand upon these concepts that constitute an end-to-end document understanding solution.

\section{Optical Character Recognition}\label{sec:ocr}
\begin{figure}[t]
    \centering
    \includegraphics[width=\textwidth]{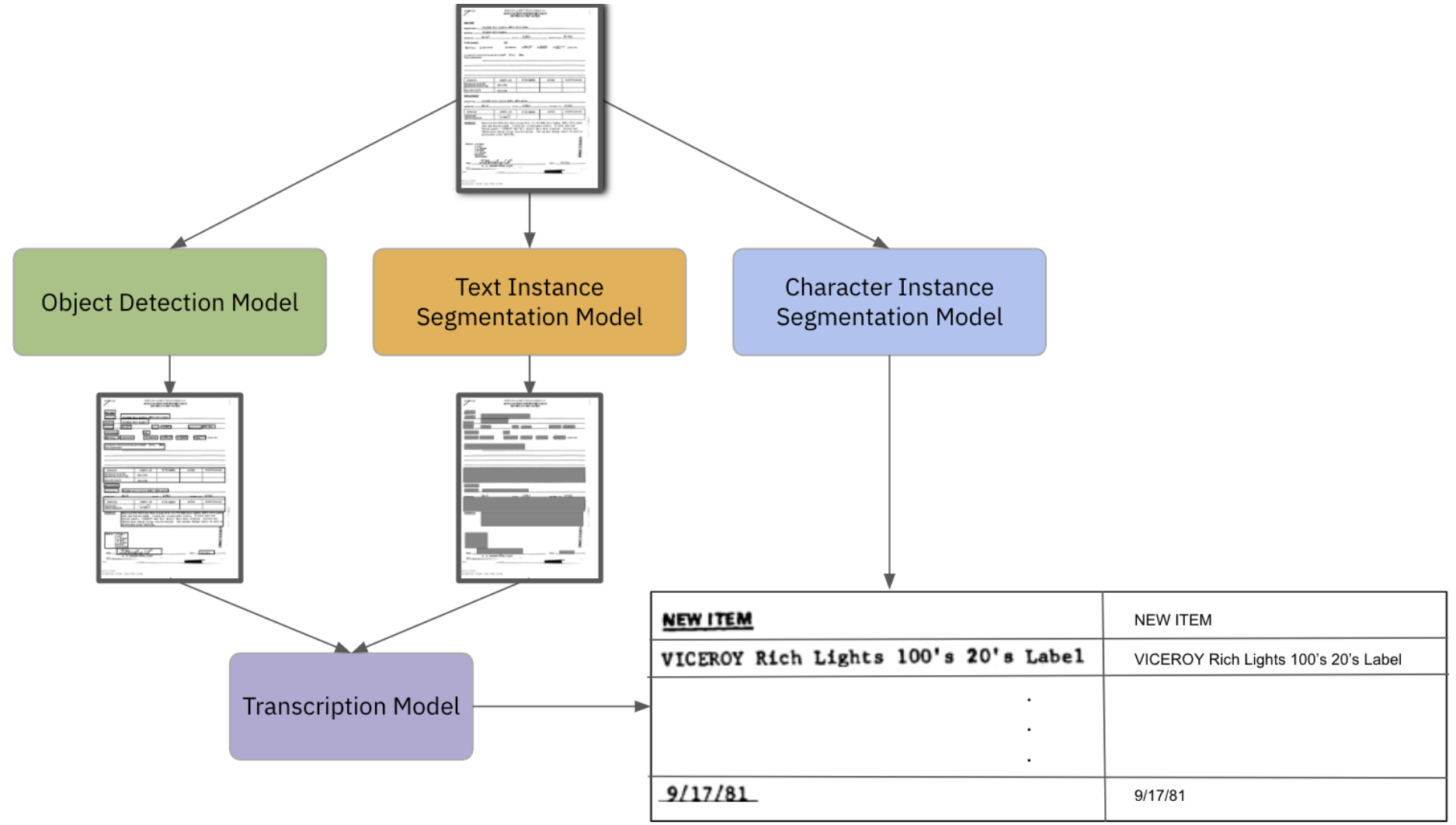}
    \caption{Here, we show the general OCR process. A document can take the left path and go through an object detection model, which outputs bounding boxes, and a transcription model that transcribes the text in each of those bounding boxes. If the document takes the middle path, the object passes through a generic text instance segmentation model that colors pixels black if they contain text and a text transcription model that transcribes the regions of text the instance segmentation model identifies. If the document takes the right path, the model goes through a character-specific instance segmentation model, which outputs which character a pixel corresponds to. All paths produce the same structured output. The document comes from FUNSD~\citep{jaume2019funsd}.}
    \label{fig:ocr-figure}
\end{figure}

OCR has two primary components: text detection and text transcription.
Generally, these two components are separate and employ different models for each task.
Below, we discuss state-of-the-art methods for each of these components and show how a document can be processed through different generic OCR systems. See Figure~\ref{fig:ocr-figure} for details.

\subsection{Text Detection}\label{sec:text-detection}

Text detection is the task of finding text present in a page or image.
The input, an image, is often represented by a three dimensional tensor, $C \times H \times W$, where $C$ is the number of channels (often three, for red, green and blue), $H$ is the height, and $W$ is the width of the image.
Text detection is a challenging problem because text comes in a variety of shapes and orientations and can often be distorted.
We explore two common ways researchers pose the text detection problem: as an object detection task and as a instance segmentation task. A text detection model must either learn to output coordinates of bounding boxes around text (object detection), or a mask, where pixels with text are marked and pixels without are not (instance segmentation).

\subsubsection{Text Detection as Object Detection}\label{sec:text-detection-object-detection}
Traditionally, text detection revolved around hand-crafting features to detect characters~\citep{matas2004robust, lowe2004distinctive}.
Advances in deep learning, especially in object detection and semantic segmentation, have led to a change in how text detection is tackled.
Using these well-performing object detectors from the traditional computer vision literature, such as the Single-Shot MultiBox Detector (SSD) and Faster R-CNN models~\citep{liu2016ssd, ren2015faster}, practitioners build efficient text detectors.

One of the first papers applying a regression-based detector for text is TextBoxes~\citep{liao2016textboxes, liao2018textboxes++}. 
They added long default boxes that have large aspect ratios to SSD, in order to adapt the object detector to text.
Several papers built on this work to make regression-based models resilient to orientations, like the Deep Matching Prior Network (DMPNet) and the Rotation-Sensitive Regression Detector (RRD)~\citep{liu2017deep,liao2018rotation}.
Other papers have a similar approach to the problem, but develop their own proposal network that is tuned towards text rather than towards natural images.
For instance, \cite{tian2016detecting} combine convolutional networks with recurrent networks using a vertical anchor mechanism in their Connectionist Text Proposal Network to improve accuracy for horizontal text.

Object detection models are generally evaluated via an intersection over union (IoU) metric and an F1 score.
The metric computes how much of a candidate bounding box overlaps with the ground truth bounding box (the intersection) divided by the total space occupied by both the candidate and ground truth bounding boxes (the union).
Next, an IoU threshold $\tau$ is chosen to determine which predicted boxes count as true positives (IoU $\geq \tau$).
The remainder are classified as false positives.
Any box that the model fails to detect is classified as a false negative.
Using those definitions, an F1 score is computed to evaluate the object detection model.

\subsubsection{Text Detection as Instance Segmentation}\label{sec:text-detection-instance-segmentation}

Text detection in documents has its own unique set of challenges: notably, the text is usually dense and documents contain a lot more text than what is usually present in natural images. To combat this density problem, text detection can be posed as an ultra-dense instance segmentation task.
Instance segmentation is the task of classifying each pixel of an image as specific, pre-defined categories.

Segmentation-based text detectors work at the pixel level to identify regions of text.
These per-pixel predictions are often used to estimate probabilities of text regions, characters, and their relationships among adjacent characters in a unified framework.
Practitioners use popular segmentation methods like Fully Convolutional Networks (FCN) to detect text~\citep{long2015fully}, improving upon object detection models, especially when text is misaligned or distorted.
Several papers build on this segmentation foundation to output word bounding areas by extracting bounding areas directly from the segmentation output~\citep{yao2016scene, he2017multi, deng2018pixellink}.
TextSnake extends this further by predicting the text region, center line, direction of text, and candidate radius from an FCN~\citep{long2018textsnake}.
These features are then combined with a striding algorithm to extract the central axis points to reconstruct the text instance.

\subsection{Word-level versus character-level}\label{sec:text-transcription}

While most papers cited above try to directly detect words or even lines of words, some papers argue that character-level detection is an easier problem than general text detection because characters are less ambiguous than text lines or words.
CRAFT uses an FCN model to output a two-dimensional Gaussian heatmap for each character~\citep{baek2019character}.
Characters that are close together are then grouped together in a rotated rectangle that has the smallest area possible to still encapsulate the set of characters.
More recently, \cite{Ye2020TextFuseNetST} combine global, word-level, and character-level features obtained using Region Proposal Networks (RPN) to great success.

Most of the models described above were mainly developed for text scene detection, but can be easily adapted to document text detection to handle difficult cases like distorted text.
We expect less distortion in documents than in natural images, but poorly scanned documents or documents with certain fonts could still pose these problems.

\subsection{Text Transcription}\label{sec:text-transcription}
Text transcription is the task of transcribing the text present in an image.
The input, an image, is often a crop corresponding to either a character, word, or sequence of words, and has dimension $C \times H' \times W'$.
A text transcription model must learn to ingest this cropped image and output a sequence of tokens belonging to some pre-specified vocabulary $V$.
$V$ often corresponds to a set of characters. For digit recognition for instance, this is the most intuitive approach~\citep{goodfellow2013multi}.
Otherwise, $V$ can also correspond to a set of words, similarly to a word-level language modeling problem~\citep{Jaderberg2014SyntheticDA}. 
In both cases, the problem can be framed as a multi-class classification problem with the number of classes equal to the size of the vocabulary $V$.

Word-level text transcription models require more data as the number of classes in the multi-class classification problem is much larger than for character-level. On one hand, predicting words instead of characters decreases the probability of making small typos (like replacing an "a" by an "o" in a word like "elephant"). On the other, limiting oneself to a word-level vocabulary means that it is not possible to transcribe words which are not part of this vocabulary. This problem doesn't exist at the character-level, as the number of characters is limited. As long as we know the language of the document, it is straightforward to build a vocabulary which contains all the possible characters. Subword units are a viable alternative~\citep{sennrich2015neural}, as they alleviate the issues present in both word and character level transcription. 

Recently the research community has moved towards using recurrent neural networks, specifically recurrent models with LSTM or GRU units on top of a convolutional image feature extractor~\citep{hochreiter1997long, cho2014learning, Wang2017GatedRC}.
To transcribe a token, two different decoding mechanisms are often used.
One is standard greedy decoding or beam search using an attention-based sequence decoder with cross entropy loss~\citep{bahdanau2014neural}, exactly like decoding with a conditional language model.
Sometimes images are poorly oriented or misaligned, reducing the effectiveness of standard sequence attention.
To overcome this,~\cite{he2018end} uses attention alignment, encoding spatial information of characters directly, while~\cite{Shi2016RobustST} use spatial attention mechanisms directly.
The second way in which transcription decoding is often done is with connectionist temporal classification (CTC) loss~\citep{graves2006connectionist}, a common loss function in speech which models repeated characters in sequence outputs well.

The majority of text transcription models borrow from advances in sequence modeling for both text and speech and often can utilize these advancements well with only minor adjustments.
As a result, practitioners seldom directly tackle this aspect relative to the other components of the document understanding task.

\subsection{End-to-end models}\label{sec:combined-models}

End-to-end approaches combine text detection and text transcription in order to improve both components jointly~\citep{li2017towards}.
For instance, if the text prediction has a very low probability, it means the detected box either did not capture the entire word or captured something that is not text.
An end-to-end approach may be very effective in this case.
Combining these two methods is fairly common and both Fast Oriented Text Spotting (FOTS) and TextSpotter with Explicit Alignment and Attention sequentially combine these models to train end-to-end~\citep{liu2018fots, he2018end}.
These approaches use shared convolutions as features to both text detection and recognition, and implement methods for complex orientations of text.
\cite{feng2019textdragon} introduce TextDragon, an end-to-end model that performs well on distorted text by utilizing a differentiable region of interest slide operator, which specializes in correcting distortions in regions of interest. 
Mask TextSpotter is another end-to-end model that combines region proposal networks for bounding boxes with text and character segmentation~\citep{liao2020mask}.
These recent works show the power of end-to-end OCR solutions in reducing errors.

Yet, having separate text detection and text recognition models offers more flexibility. First, the two models can be trained separately. In the case where only a small dataset is available to train the whole OCR module, but a lot of text recognition data is easily accessible, it makes sense to leverage this big amount of data in the training of the recognition model. Moreover, with two separate models, it is easy to compute two separate sets of metrics and have a more complete understanding of where the bottleneck might be.

Hence, both two-model and end-to-end approaches are viable. Whether one is better than the other mainly depends on the data available and what one wants to achieve.

\subsection{Datasets for Text Detection \& Transcription}\label{sec:text-detection-datasets}
Most of the literature revolves around scene text detection, rather than document text detection, and report results on those datasets.
Some of the major ones are ICDAR~\citep{karatzas2013icdar, karatzas2015icdar}, Total-Text~\citep{ch2017total}, CTW1500~\citep{yuliang2017detecting}, and SynthText~\citep{gupta2016synthetic}.

\cite{jaume2019funsd} present FUNSD, a dataset for text detection, transcription, and document understanding with 199 fully annotated forms comprising of 31k word level bounding boxes. Another recent document understanding dataset comes from the ICDAR 2019 Robust Reading Challenge on Scanned Receipts OCR and Information Extraction (SROIE). It contains 1000 whole scanned receipt images, with line-level annotations for text detection/transcription, and labels for Key Information Extraction. The website contains a ranking of the solutions proposed to address this problem. As solutions are still posted after the end of the competition, it is a good way to keep track of the most recent methods.

\section{Document Layout Analysis}\label{sec:layout-analysis}

Document layout analysis is the process of locating and categorizing regions of interest on a picture or scanned image of a page.
Broadly, most approaches can be distilled into page segmentation and logical structural analysis~\citep{layoutAnalysisSurvey2019, okun1999page}.
Page segmentation methods focus on appearance and use visual cues to partition pages into distinct regions; the most common are text, figures, images, and tables. 
In contrast, logical structural analysis focuses on providing finer-grained semantic classifications for these regions, i.e. identifying a region of text that is a paragraph and distinguishing that from a caption or document title.

Research in methods for document layout analysis has a long history, both in academia and industry.~\footnote{The first ISO standard that defined aspects of modern-day document layout analysis was drafted four decades ago: ISO 8613-1:1989}
From the first pioneering heuristic approaches~\citep{layout201771, okun1999page, liang1997uw}, to multi-stage classical machine learning systems~\citep{8480095, wei2013evaluation, eskenazi2017comprehensive}, the evolution of document layout analysis methods is now dominated by end-to-end differentiable methods~\citep{yang2017learning, layoutAnalysisSurvey2019, Ares_Oliveira_2018, Agarwal2020CDeCNetCD, Monnier2020docExtractorAO, pramanik2020towards}.

\subsection{Instance Segmentation for Layout Analysis}\label{sec:instance-segmentation}

When applied to the problem of layout analysis in business documents, instance segmentation methods predict per-pixel labels to categorize regions of interest.
Such methods are flexible and easily adapt to the courser-grained task of page segmentation or the more-specific task of logical structural analysis.

\begin{figure}[t]
    \label{fig:layout-analysis}
    \centering
    \includegraphics[width=\textwidth]{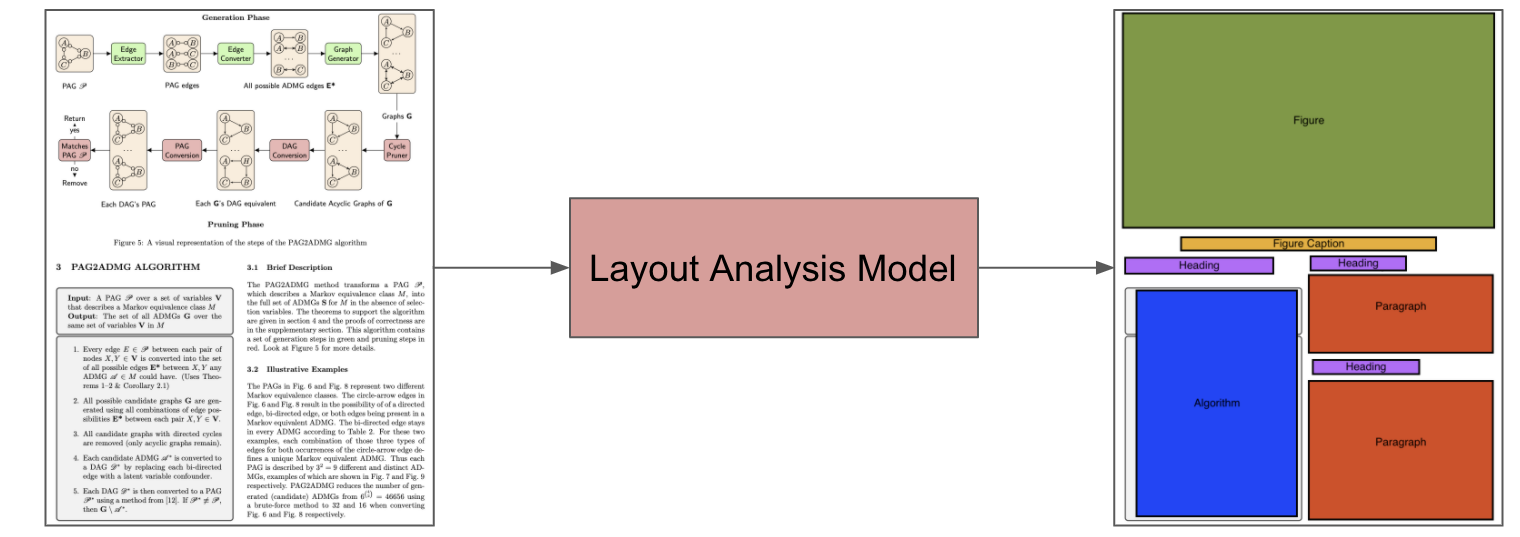}
    \caption{A document is passed through a generic layout analysis model, resulting in a layout segmentation mask with the following classes: figure (green), figure caption (orange), heading (purple), paragraph (red), and algorithm (blue). The document has been reproduced with permission~\citep{subramani2016pag2admg}.}
\end{figure}

In \cite{yang2017learning}, the authors describe an end-to-end neural network that combines both text and visual features in a encoder-decoder architecture that also incorporates an unsupervised pretraining network.
During inference, their approach uses a downsampling cascade of pooling layers to encode visual information, which is fed into a symmetrical upsampling cascade for decoding. 
At each cascade level, the produced encoding is also directly passed into the respective decoding block, concatenating the down- and up-sampled representations.
This architecture ensures that visual feature information at different levels of resolution is considered during the encoding and decoding process~\citep{imagePyramid}.
For the final decoding layer, localized text embeddings are supplied alongside the computed visual representation.

This U-Net inspired encoding-decoding architecture has been adopted for document layout analysis in several different approaches~\citep{ronneberger2015u}.
The method in \cite{Ares_Oliveira_2018}, and later extended by \cite{barman2020combining} via additional text embeddings, use convolution maxpooling layers with large filter sizes to feed the document image through a ResNet bottleneck~\citep{he2015deep}.
The representation is then processed by bilinear upsampling layers and smaller 1x1 and 3x3 convolution layers.
Both works are used to perform layout analysis on historical documents and newspapers from multiple European languages, respectively.
In \cite{lee2019page}, the authors combine the U-Net architecture pattern with trainable multiplication layers.
This layer type is specialized for extracting co-occurrence texture features from the network's convolution feature maps, which are effective for locating regions that have periodically repeating information, such as tables.

\subsection{Addressing Data Scarcity and Alternative Approaches}

Obtaining high quality training data for layout analysis is a labor intensive task that requires both mechanical precision and an understanding of the document's contents.
As a consequence of the difficulties in layout annotation of documents from brand new domains, several approaches exist to either leverage structure in unlabeled data or use well-defined rule sets to generate synthetic labeled documents to further improve generalizability and performance of document layout analysis systems.

Masked language models such as BERT and RoBERTa have shown effective empirical performance on many downstream NLP tasks~\citep{Devlin2019BERTPO, liu2019roberta}.
Inspired by the pretraining strategy in BERT and RoBERTa, \cite{Xu2020} define a Masked Visual-Language Model, which randomly masks input tokens and uses the model to predict the masked tokens. Unlike BERT, their method provides the 2-D positional embedding of the token during this masked prediction task, which enables the model to combine both semantic and spatial relationships between textual elements. Mentioned earlier in section~\ref{sec:instance-segmentation}, \cite{yang2017learning} introduce an auxiliary document image reconstruction task in their broader instance segmentation-based network. During training, this auxiliary module uses a separate upsampling decoder that, without the aid of skip connections, predicts the original pixel values from the encoded representation.

While pretraining lets practitioners gain more value from their unlabeled documents, this technique alone is not always sufficient to effectively surmount data scarcity concerns. Relying on the intuition that many business and academic documents have repeated patterns in both content as well as page-level organization, several approaches have emerged to manufacture synthetic, labeled data in order to provide data suitable for a pretraining-like routine \citep{zhong2019publaynet}.
In \cite{Monnier2020docExtractorAO}, the authors propose a three-stage method for synthesizing new labeled documents.  First, they generate the document by randomly choosing the a document background from a set of nearly 200 known document backgrounds. Second, they use a grid based layout method to define both individual document element content and their respective sizes. Third, their process introduces corruptions, such as Gaussian blur and random image crops. This modular, rule-based synthetic document generation approach creates a heterogeneous dataset to make pretraining of layout analysis models more robust.

Alternatively, instead of defining rules to generate a heterogeneous set of documents, several synthesizing procedures take cues from data augmentation methods.
\cite{8270127} and \cite{Journet_2017} describe general-purpose toolkits that use an existing set of labeled documents to introduce deformations and perturbations in source images.
Importantly, such changes to the training data are balanced so as to preserve the original semantic content while still exposing the model training to realistic errors that it must account for during inference on unseen data.

\subsection{Datasets for Layout Analysis}

Recently, there has been a deluge of datasets specifically targeting the document layout analysis problem. The International Conference on Document Analysis and Recognition (ICDAR) has produced several datasets from their various annual competitions; the most recent from 2017 and 2019 provide gold-standard data for document layout analysis and other document processing tasks \citep{icdar2017, Clausner2019ICDAR2019CO, Huang2019ICDAR2019CO, Gao2019ICDAR2C}.

On the larger side, DocBank is a a collection of half a million document pages with token-level annotations suitable for training and evaluating document layout analysis systems \citep{li2020docbank}. The authors constructed this dataset using weak supervision \citep{hoffmann2011}, matching data from the LaTeX source of known PDFs to form annotations. Similarly, \cite{zhong2019publaynet} created PubLayNet by automatically matching XML content representations for over one million PDFs on PubMed Central™, consiting of approximately 360 thousand document images. While not full document layout, \cite{Zhong2019ImagebasedTR} have created PubTabNet from PubMed Central as well. Their data consists of 568 thousand table images alongside an HTML representations of content.

\section{Information Extraction}\label{sec:information-extraction} 

The goal of information extraction for document understanding is to take documents that may have diverse layouts and extract information into a structured format.
Examples include receipt understanding to identify item names,  quantities, and prices and form understanding to identify different key-value pairs.
Document extraction of information by humans goes beyond simply reading text on a page as it is often necessary to learn page layouts for complete understanding.
As such, recent enhancements have extended text encoding strategies for documents by additionally encoding structural and visual information of text in a variety of ways.

\subsection{2D Positional Embeddings}
Multiple sequence tagging approaches have been proposed which augment current named entity recognition (NER) methods by embedding attributes of 2D bounding boxes and merging them with text embeddings to create models which are simultaneously aware of both context and spatial positioning when extracting information.
\cite{Xu2020} embeds the pair of $x,y$ coordinates that define a bounding box using two different embedding tables and pretrain a masked language model (LM).
During pretraining, text is randomly masked but the 2D positional embeddings are retained.
This model can then be fine-tuned on a downstream task.
Alternatively, the bounding box coordinates can also be embedded using $sin$ and $cos$ functions like positional encoding methods~\citep{vaswani2017attention,hwang2020spatial}.
Other features can also be embedded such as the line or sequence number~\citep{hwang2019postocr}.
In this scenario, the document is preprocessed to assign a line number to each individual token. Each token is then ordered from left to right and given a sequential position. Finally, both the line and sequential positions are embedded.

While these strategies have seen success, relying solely on the line number or bounding box coordinates can be misleading when the document has been scanned on an uneven surface, leading to curved text.
Additionally, bounding box based embeddings still miss critical visual information such as typographical emphases (bold, italics) and images such as logos.
To overcome these, a crop of the image corresponding to the token of interest can be embedded using a Faster R-CNN model to create token image embeddings which are combined with the 2D positional embeddings~\citep{Xu2020}.

\subsection{Image Embeddings}

Information extraction for documents can also be framed as a computer vision challenge wherein the goal of the model is to semantically segment information or regress bounding boxes over the areas of interest.
This strategy helps preserve the 2D layout of the document and allows models to take advantage of 2D correlations.
While it is theoretically possible to learn strictly from the document image, directly embedding textual information into the image simplifies the task for models to understand the 2D textual relationships.
In these cases, an encoding function is applied onto a proposed textual level (i.e. character, token, word) to create individual embedding vectors.
These vectors are  transposed into each pixel that comprises the bounding box corresponding to the embedded text, ultimately creating an image of $W \times H \times D$ where $W$ is the width, $H$ is the height, and $D$ is the embedding dimension.
Proposed variants are listed as following:
\begin{enumerate}
    \item \textit{CharGrid} embeds characters with a one-hot encoding into the image~\citep{Katti2018ChargridTU} 
    \item \textit{WordGrid} embeds individual words using word2vec or FastText~\citep{kerroumi2020visualwordgrid} 
    \item \textit{BERTgrid} finetunes BERT on task-specific documents and is used obtain contextual wordpiece vectors~\citep{Denk2019BERTgridCE}
    \item \textit{C+BERTgrid}, combines context-specific and character vectors~\citep{Denk2019BERTgridCE}
\end{enumerate}

When comparing the grid methods, C+BERTgrid has shown the best performance, likely due to its contextualized word vectors combined with a degree of resiliency to OCR errors. 
\cite{zhao2019cutie} proposes an alternative approach to directly apply text embeddings to the image.
A grid is projected on top of the image and a mapping function assigns each token to a unique cell in the grid.
Models then learn to assign each cell in the grid to a class.
This method significantly reduces the dimensionality due to its grid system, while still retaining the majority of the 2D spatial relationships. 

\subsection{Documents as Graphs}

Unstructured text on documents can also be represented as graph networks, where the nodes in a graph represent different textual segments.
Two nodes are connected with an edge if they are cardinally adjacent to each other, allowing the relationship between words to be modeled directly~\citep{qian2019graphie}.
An encoder such as a BiLSTM encodes text segments into nodes~\citep{qian2019graphie}.
Edges can be represented as a binary adjacency matrix or a richer matrix, encoding additional visual information such as the distance between segments or shape of the source and target nodes~\citep{liu2019graph}.
A graph convolutional network is then applied at different receptive fields in a similar fashion to dilated convolutions~\citep{yu2015multi} to ensure that both local and global information can be learned~\citep{qian2019graphie}.
After this, the representation is passed to a sequence tagging decoder.

Documents can also be represented as a directed graph and a spatial dependency parser~\citep{hwang2020spatial}. 
In this representation, nodes are represented by textual segments, but field nodes denoting the node type are used to initialize each DAG.
In addition, two kinds of edges are defined:
\begin{enumerate}
    \item Edges that group together segments belonging to the same category (STORENAME $\rightarrow$ Peet's $\rightarrow$ Coffee; a field node followed by two nodes representing a store name)
    \item Edges that connect relationships between different groups (Peet's $\rightarrow$ 94107; a zipcode).
\end{enumerate}
A transformer with an additional 2D positional embedding is used to spatially encode the text.
After this, the task becomes to predict the relationship matrix for each edge type.
This method can represent arbitrarily deep hierarchies and can be applied towards complicated document layouts.

\subsection{Tables}\label{sec:table-extraction}

Tabular data extraction remains a challenging aspect of information extraction due to their wide variety of formats and complex hierarchies. Table datasets typically have multiple tasks to perform~\citep{Shahab2010AnOA, 6628853, Gao2019ICDAR2C, Zhong2019ImagebasedTR}. The first task is table detection which involves localizing the bounding box containing the table(s) inside the document. The next task is table structure recognition, which requires extracting the row, column, and cell information into a common format. This can be taken one step further to table recognition, which requires understanding both the structural information as well as the content by classifying cells within the table itself~\citep{Zhong2019ImagebasedTR}. As textual and visual features are equally important to properly extracting and understanding tables, many diverse methods have been proposed to perform this task.

One such proposal named TableSense performs both table detection and structure recognition~\citep{Dong2019TableSenseST}. 
TableSense uses a three stage approach: cell featurization, object detection with convolutional models, and an uncertainty-based active learning sampling mechanism. TableSense's proposed architecture for table detection performs significantly better than traditional methods in computer vision such as YOLO-v3 or Mask R-CNN~\citep{Redmon2018YOLOv3AI, he2017mask}.
Since this approach does not work well for general spreadsheets, ~\cite{dong2019semantic} extend upon the previous work by using a multitask framework to jointly learn table regions, structural components of spreadsheets, and cell types.
They add an additional stage, which leverages language models to learn the semantic contents of table cells in order to flatten complex tables into a single standard format.

\cite{wang2020structureaware} propose TUTA, which focuses on understanding the content within tables after the structure has been determined. The authors present three new objectives for language model pretraining for table understanding by using tree-based transformers. The objectives introduced for pretraining are designed to help the model understand tables at the token, cell, and table level. The authors mask a proportion of tokens depending on the table cell for the model to predict, randomly mask particular cell headers for the model to predict the header string based on its location, and provide the table with context such as table titles or descriptions that may or may not be associated for the model to identify which contextual elements are positively associated with the table. The transformer architecture is modified to reduce distractions from attention by limiting the attention connections to items based on a cell's  hierarchical distance to another cell. Fine-tuning TUTA has demonstrated state of the art performance on multiple datasets for cell type classification.

\section{Conclusion}
Document understanding is a hot topic in industry and has immense monetary value.
Most documents are private data corresponding to private contracts, invoices, and records.
As a result, openly available datasets are hard to come by and have not been a focus for academia with respect to other application areas.
The academic literature on methodologies to tackle document understanding is similarly sparse relative to areas with an abundance of publicly available data such as image classification and translation.
However, the most effective approaches for document understanding make use of recent advancements in deep neural network modeling.
End-to-end document understanding is achievable by creating an integrated system that performs layout analysis, optical character recognition, and domain-specific information extraction.
In this survey, we attempt to consolidate and organize the methodologies which are present in literature in order to be a jumping-off point for scholars and practitioners alike who want to explore document understanding.

\bibliography{references}

\end{document}